\documentclass[sigconf]{acmart}
\usepackage{multirow} 
\AtBeginDocument{%
  \providecommand\BibTeX{{%
    \normalfont B\kern-0.5em{\scshape i\kern-0.25em b}\kern-0.8em\TeX}}}

\setcopyright{acmcopyright}
\copyrightyear{2022}
\acmYear{2022}
\acmDOI{10.1145/3528233.3530744}

\acmConference[SIGGRAPH '22 Conference Proceedings]{Special Interest Group on Computer Graphics and Interactive Techniques Conference Proceedings}{August 7--11, 2022}{Vancouver, BC, Canada}
\acmBooktitle{Special Interest Group on Computer Graphics and Interactive Techniques Conference Proceedings (SIGGRAPH '22 Conference Proceedings), August 7--11, 2022, Vancouver, BC, Canada}
\acmPrice{15.00}
\acmISBN{978-1-4503-9337-9/22/08}

\begin{document}

\title[ImLoveNet]{ImLoveNet: Misaligned Image-supported Registration Network for Low-overlap Point Cloud Pairs}

\author{Honghua Chen}
\email{chenhonghuacn@gmail.com}
\orcid{0000-0001-7473-1146}
\affiliation{%
  \institution{Nanjing University of Aeronautics and Astronautics}
  \streetaddress{No.29, Yudao Street}
  \city{Nanjing}
  \country{China}
}

\author{Zeyong Wei}
\email{weizeyong1@gmail.com}
\orcid{0000-0003-2684-9769}
\affiliation{%
  \institution{Nanjing University of Aeronautics and Astronautics}
  \streetaddress{No.29, Yudao Street}
  \city{Nanjing}
  \country{China}}

\author{Yabin Xu}
\email{yabinxu007@gmail.com}
\orcid{0000-0002-1919-8673}
\affiliation{%
  \institution{Nanjing University of Aeronautics and Astronautics}
  \streetaddress{No.29, Yudao Street}
  \city{Nanjing}
  \country{China}}

\author{Mingqiang Wei}
\email{mingqiang.wei@gmail.com}
\orcid{0000-0003-0429-490X}
\affiliation{%
  \institution{Nanjing University of Aeronautics and Astronautics}
  \streetaddress{No.29, Yudao Street}
  \city{Nanjing}
  \country{China}}

\author{Jun Wang}
\email{wjun@nuaa.edu.cn}
\orcid{0000-0001-9223-2615}
\affiliation{%
  \institution{Nanjing University of Aeronautics and Astronautics}
  \streetaddress{No.29, Yudao Street}
  \city{Nanjing}
  \country{China}}

\renewcommand{\shortauthors}{Chen et al.}

\begin{abstract}
Low-overlap regions between paired point clouds make the captured features very low-confidence, leading cutting edge models to point cloud registration with poor quality. Beyond the traditional wisdom, we raise an intriguing question: Is it possible to exploit an intermediate yet misaligned image between two low-overlap point clouds to enhance the performance of cutting-edge registration models? To answer it, we propose a misaligned image supported registration network for low-overlap point cloud pairs, dubbed ImLoveNet. ImLoveNet first learns triple deep features across different modalities and then exports these features to a two-stage classifier, for progressively obtaining the high-confidence overlap region between the two point clouds. Therefore, soft correspondences are well established on the predicted overlap region, resulting in accurate rigid transformations for registration. ImLoveNet is simple to implement yet effective, since 1) the misaligned image provides clearer overlap information for the two low-overlap point clouds to better locate overlap parts; 2) it contains certain geometry knowledge to extract better deep features; and 3) it does not require the extrinsic parameters of the imaging device with respect to the reference frame of the 3D point cloud. Extensive qualitative and quantitative evaluations on different kinds of benchmarks demonstrate the effectiveness and superiority of our ImLoveNet over state-of-the-art approaches.
\end{abstract}



\begin{teaserfigure}
  \includegraphics[width=\textwidth]{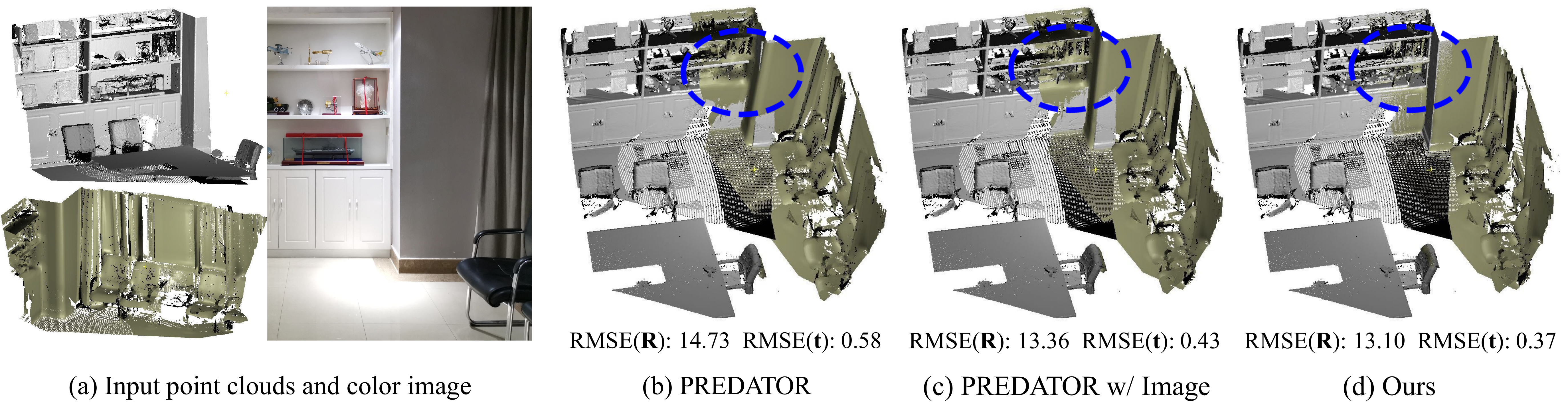}
  \caption{Our method is able to well align low-overlap point cloud pairs with the assistance of a misaligned intermediate color image. Such a new configuration for the point cloud registration is also useful to enhance the performance of other registration model, like PREDATOR \cite{huang2021predator}.}
  \label{fig:teaser}
\end{teaserfigure}

\maketitle

\section{Introduction}

With the rapid development of new 3D acquisition technologies, 3D sensors are becoming increasingly available and affordable, including various types of 3D laser scanners (or LiDAR), and RGB-D cameras (such as Microsoft Kinect, Intel RealSense, and Apple Truth Depth Camera). 
This benefits to acquire more reliable 3D information, enabling better understanding of surrounding large-scale environment for machines. 
Consequently, these sensors greatly increase the expectations on the performance of point cloud registration.
Point cloud registration aims at finding a rigid transformation to align a pair of point clouds. 
It has wide applications from low-level 3D reconstruction to higher-level scene analysis or applied robotics. 

\begin{figure}
	\centering
	\includegraphics[width=0.5\textwidth]{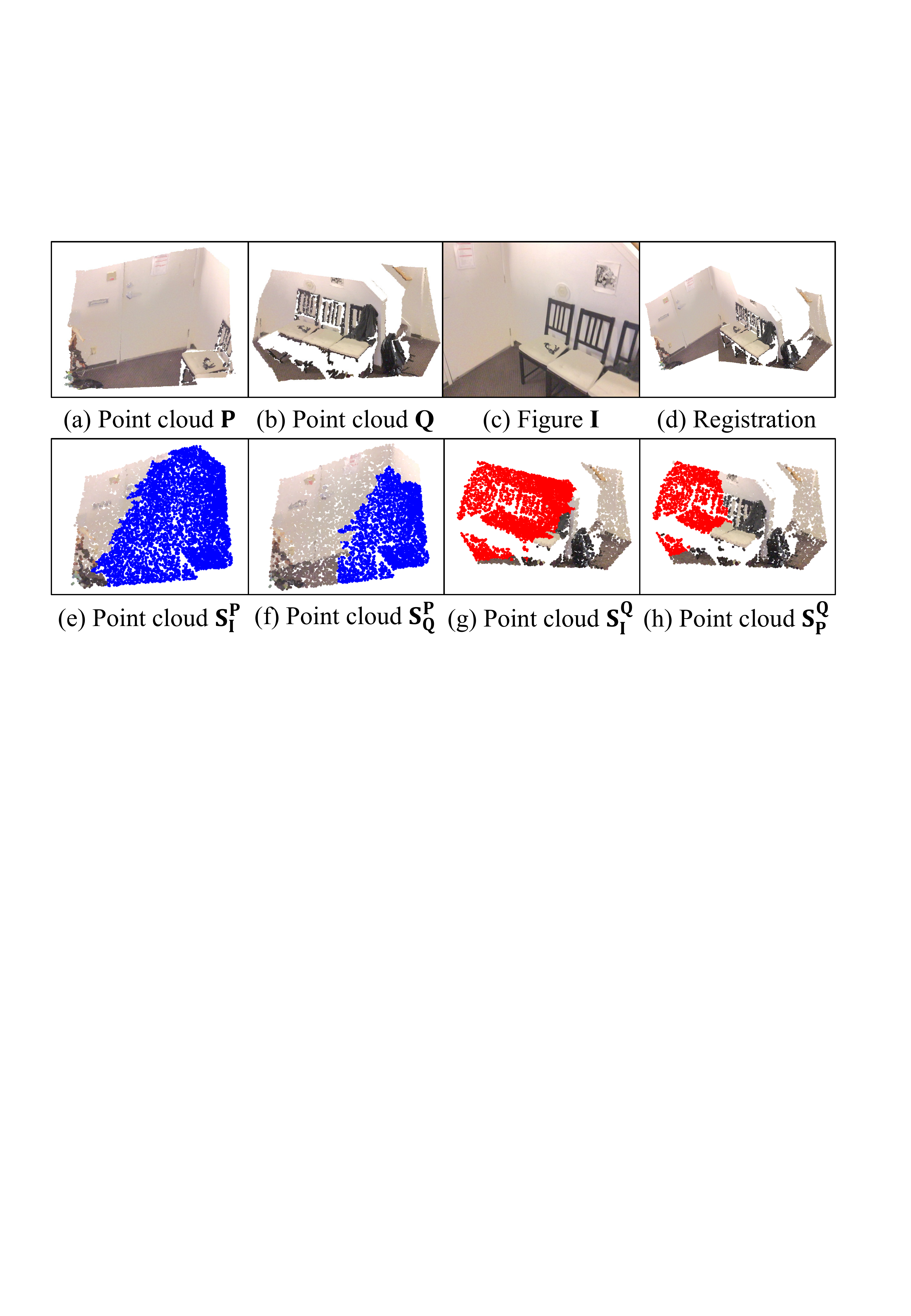}
	\caption{Illustration of low-overlap point cloud registration (overlap ratio $<20\%$) with an intermediate image. (a) and (b) are two point clouds to register. (c) is an intermediate color image. (d) is the registration result of our ImLoveNet. $\mathbf{S}^\mathbf{P}_\mathbf{I}$ (blue points) and  $\mathbf{S}^\mathbf{Q}_\mathbf{I}$ (red points) denote the points which can be projected to the image space.  $\mathbf{S}^\mathbf{P}_\mathbf{Q}$ (blue points) and $\mathbf{S}^\mathbf{Q}_\mathbf{P}$ (red points) are overlap region between the two point clouds. We append texture color on each point only for better visualization.}
	\label{fig:example}
\end{figure}

Recent work has made substantial and impressive progress in automatic point cloud registration with deep learning, like \cite{xu2021omnet,huang2021predator,pais20203dregnet,choy2020deep,wang2019deep,wang2019prnet}. However, when the overlap-region ratio between two point clouds falls below $30\%$ or less, the registration performance of these methods deteriorates rapidly  \cite{xu2021omnet,huang2021predator}. This is because it is difficult to perceive reliable corresponding information from the entire point cloud pairs with limited overlap area, which is very common in many practical scenarios. For example, data acquisition for large-scale objects is often time-consuming (Completely scanning an object of size $10m\times5m$ consumes over 12 hours, by a structured-light scanner), so practitioners aim for a low number of scans with only the necessary overlap. Also, it may be difficult to ensure a high overlap ratio for a moving scanner, when suffering from occlusions, missing frames, or large deviation of the scanning angle of view.

One interesting thing is that if an intermediate color image depicting the rough overlapping region is given, it will be easier for a human operator to register two low-overlap scans. Intuitively, with the help of the intermediate image, i) overlapping information is significantly clearer and ii) 2D image can provide certain underlying 3D-aware geometry information, even though the information is cross-modality. 
For example, we have witnessed many impressive 3D-related tasks on a single image \cite{saha2021learning, wang2020sdc, qi2018geonet}. Such kind of 3D-aware information is consistent with the real 3D feature from the point cloud to a certain extent. In addition, an intermediate image between two point clouds is easy to obtain, and we do not need to know extrinsic parameters of the imaging device with respect to the reference frame of the 3D point cloud, namely a misaligned intermediate image. All these encourage us to employ such an auxiliary image to enhance pairwise 3D point cloud registration.

To this end, we propose a misaligned image-supported registration network for low-overlap point cloud pairs, dubbed ImLoveNet, as illustrated in the example shown in Figure \ref{fig:example}. In order to fully utilize both color image and point cloud information, our network learns triple deep features, composed of the 3D feature for the point cloud, and 2D feature and simulated 3D feature for image. These learned features are then fused and applied to progressively detect the intersecting area between the input two point clouds. Finally, soft correspondences can be well established on the predicted overlap region, which leads to the quality rigid transformation parameters for final registration. Experiments and detailed analysis show that our approach achieves state-of-the-art performance (see from Figure. \ref{fig:teaser}) compared with previous algorithms. 

Our main contributions are three-fold:

$\bullet$ We design a new point cloud registration network by collaborating cross-modality information, which shows clear improvements over the state-of-the-art methods.

$\bullet$ We extract triple features from the 2D domain, 3D domain, and mimicked 3D domain, and fuse them with attention modules, which can output more reliable features as an input of the following classification network.

$\bullet$ We propose a two-stage classifier, which can progressively locate the overlapping regions among three inputs.


\begin{figure*} %
\setlength{\abovecaptionskip}{0.1cm}
	\centering
	\includegraphics[width=\textwidth]{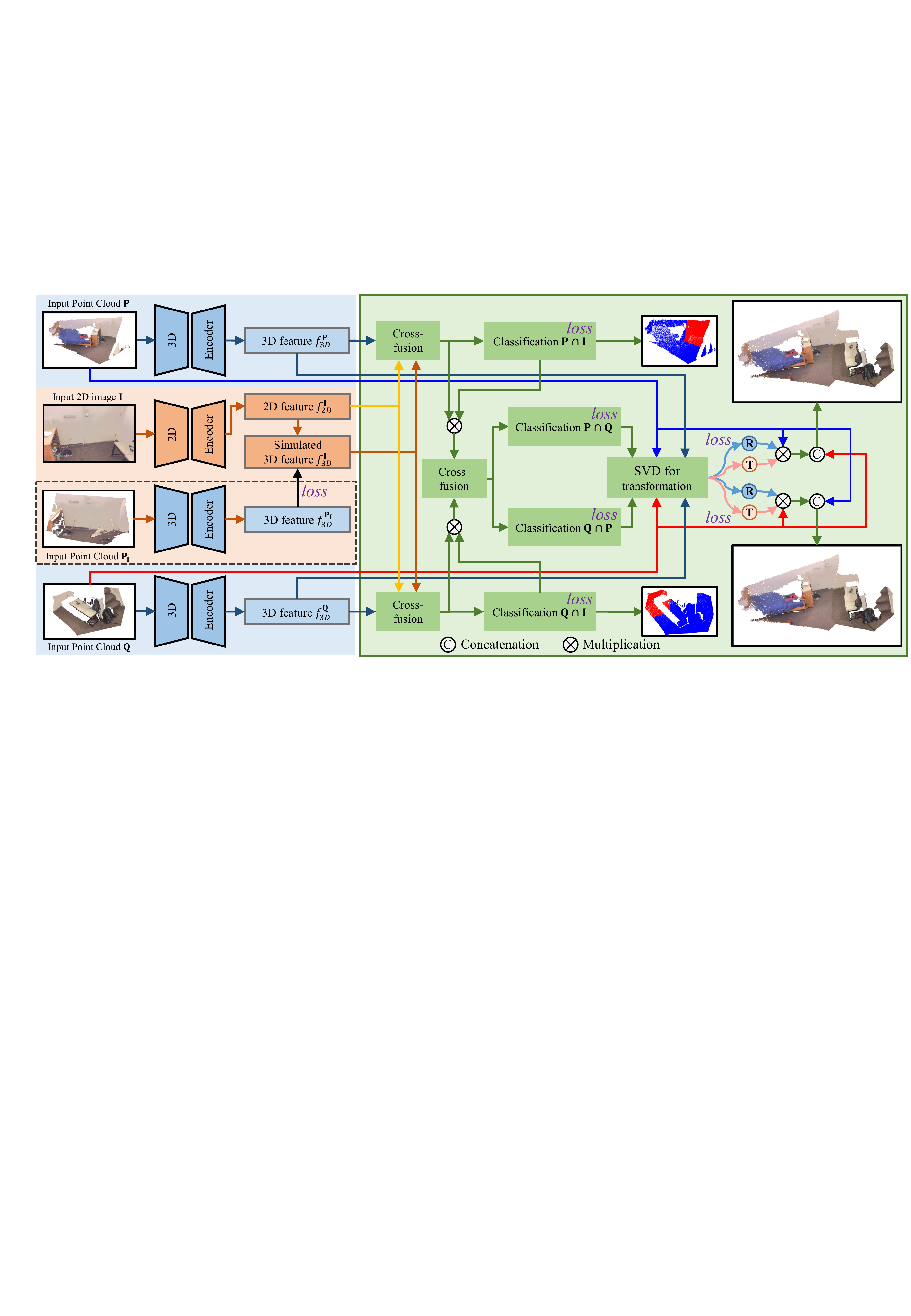}
	\caption{Network architecture of ImLoveNet. Note that we append texture colors on each point only for better visualization. We do not use the point cloud color information during both training and testing stages. Besides, $\mathbf{P}_{\mathbf{I}}$ is only used during the training stage.}
	\label{fig:architecture}
\end{figure*}
\label{sec:intro}

\section{Related work}
We start this section by reviewing the feature-based point cloud registration methods to newer end-to-end point cloud registration algorithms. Finally, we briefly cover recent advances in using cross-modality information to guide feature extraction and matching.

\subsection{3D Features for Point Cloud Registration}
Traditional methods tend to use hand-crafted 3D features that characterize the local geometry for point cloud registration, such as FPFH~\cite{rusu2009fast}, SHOT~\cite{TombariCombined2011}, or PPF~\cite{Drost2012}. Although lacking robustness in cluttered or occluded scenes, they have been widely employed in downstream applications owing to their generality across different datasets \cite{guo2014performance}. More recently, with the advances in deep neural networks, there is also a growing trend that utilizes learned 3D descriptors in point cloud registration. For instance, the pioneering work 3DMatch~\cite{zeng20173dmatch} employed a siamese deep learning architecture to extract local 3D descriptors. PPFNet \cite{deng2018ppfnet} and PPF-FoldNet \cite{deng2018ppf} proposed to combine PointNet \cite{qi2017pointnet} and PPF to extract descriptors that are aware of the global context. To improve the robustness against noise and voxelization, \cite{gojcic2019perfect} proposed to learn 3D descriptors based on a voxelized smoothed density value (SDV) representation. D3Feat \cite{bai2020d3feat} proposed a joint learning of keypoint detector and descriptor, which provides descriptors and keypoint scores for all points with extra cost during inference. \cite{wang2021you} devised a new descriptor that simultaneously has rotation invariance and rotation equivalence. Although promising results have been achieved by these methods, limitations still occur on low-overlap regions due to their locality and extracted single-modality information on point clouds.

\subsection{End-to-End Point Cloud Registration}
Apart from introducing learned keypoints and descriptors, methods~\cite{aoki2019pointnetlk, qi2017pointnet, wang2019deep, wang2019dynamic} have also been proposed to embed the differentiable pose optimization to the registration pipeline to form an end-to-end framework. \cite{Avetisyan2019} formulated a differentiable Procrustes alignment paired with a symmetry-aware dense object correspondence prediction to align CAD models to RGB-D scans. PointNetLK~\cite{aoki2019pointnetlk} designed a Lucas/Kanade like optimization algorithm that tailored to a PointNet-based~\cite{qi2017pointnet} descriptor to estimate the relative transformation in an iterative manner. DCP~\cite{wang2019deep} utilized a DGCNN network for correspondence matching, and used a differentiable SVD module for transformation estimation. PRNet~\cite{wang2019prnet} extended DCP by including a keypoint detection step and allowed for aligning partially overlapping point clouds without the need for strict one-to-one correspondence. RPM-Net \cite{yew2020rpm} used the differentiable Sinkhorn layer and annealing to get soft assignments of point correspondences from hybrid features. Later, \cite{huang2020feature} proposed FMR, which achieved pleasing results by constraining the similarity between point cloud pairs. \cite{bai2021pointdsc} designed a novel deep neural network that explicitly incorporates spatial consistency for pruning outlier correspondences. To alleviate the low-overlap registration problem, PREDATOR~\cite{huang2021predator} and OMNet~\cite{xu2021omnet} were both designed to focus more on learning the low-overlap regions. An outlier filtering network is embedded into a learned feature descriptor~\cite{choy2020deep,gojcic2020learning} to imply the weights of the correspondence in the Kabsch algorithm. At last, \cite{yan2021consistent} proposed to solve the tele-registration problem, by combining the registration and completion tasks in a way that reinforces each other.

\subsection{Cross-modality Feature Extraction and Fusion}
Recently, several algorithm have been proposed to leverage multiple sources from different channels (i.e. geometry and color information) to enhance the content of feature extraction for subsequent tasks. 3D-to-2D Distillation~\cite{liu20213d} uses an additional 3D network in the training phase to leverage 3D features to complements the RGB inputs for 2D feature extraction. Pri3D~\cite{hou2021pri3d} tried to imbue image-based perception with learned view-invariant, geometry-aware representations based on multi-view RGB-D data for 2D downstream tasks. To overcome the difficulty of cross-modality feature association, DeepI2P~\cite{li2021deepi2p} designed a neural network to covert the registration problem into a classification and inverse camera projection optimization problem. \cite{xu2021image2point} investigated the potential to transfer a pretrained 2D ConvNet to point cloud model for 3D point-cloud classification or segmentation. However, existing works tend to utilize cross-modality for enhanced segmentation or 2D-to-3D matching. In contrast, our work is the first to explore the possibility to leverage geometry and color information for point cloud registration with low-overlap region, by fusing cross-modality features.
\label{sec:related}

\section{Method}
\begin{figure*}[tb] %
	\centering
	\includegraphics[width=\textwidth]{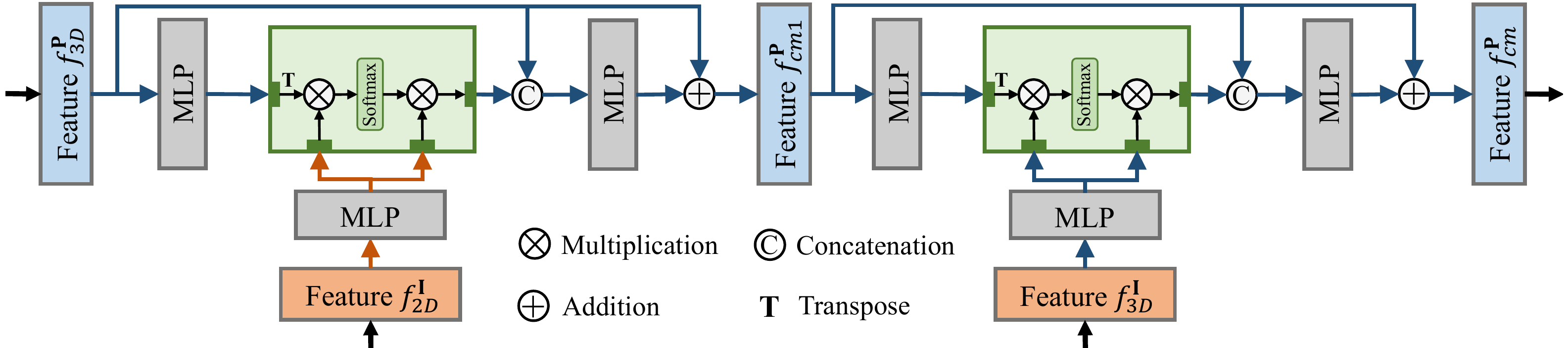}
	\caption{An illustration of cross feature fusion module.}
	\label{fig:cm_module2}
\end{figure*}

\subsection{Problem description and overview} 
We denote $\mathbf{P}=\left\{\mathbf{p}_{i} \in \mathbb{R}^{3} \mid i=1 . . . N\right\}$ and $\mathbf{Q}=\left\{\mathbf{q}_{i} \in \mathbb{R}^{3} \mid i=1 . . . M\right\}$ as the two point clouds to be registered. $\mathbf{I} \in \mathbb{R}^{W \times H \times 3}$ is the intermediate color image between $\mathbf{P}$ and $\mathbf{Q}$, and $\mathbf{P}_{\mathbf{I}}$ is the corresponding point cloud of $\mathbf{I}$, which is only used during the training stage. The overall goal is to locate the overlap region of paired point clouds and recover the rigid transformation parameters, $\mathbf{R}$ and $\mathbf{t}$. Figure \ref{fig:architecture} illustrates the overall architecture of ImLoveNet, which can be decomposed into following four main steps:

$\bullet$ Triple features extraction: extracting 3D point cloud features $f_{3D}^{\mathbf{P}} \in \mathbb{R}^{N \times C}$, $f_{3D}^{\mathbf{Q}} \in \mathbb{R}^{M \times C}$ for $\mathbf{P}$ and $\mathbf{Q}$, and the image feature $f_{2D}^{\mathbf{I}} \in \mathbb{R}^{W_1 \times H_1 \times C}$ for $\mathbf{I}$; in particular, a mimicked 3D feature $f_{3D}^{\mathbf{I}} \in  \mathbb{R}^{W_1 \times H_1 \times C}$ for $\mathbf{I}$ is additionally generated with the assistance of its paired point cloud $\mathbf{P}_{\mathbf{I}}$ (Sec. \ref{sec:fe});

$\bullet$ Cross-modality feature fusion: effectively combining the extracted triple features, e.g., $f_{3D}^{\mathbf{P}}$ (or $f_{3D}^{\mathbf{Q}}$),  image feature $f_{2D}^{\mathbf{I}}$, and simulated feature $f_{3D}^{\mathbf{I}}$ together (Sec. \ref{sec:ff});

$\bullet$ Two-stage classification: fed with the hybrid feature, detecting the points located in the camera frustum of $\mathbf{I}$ from $\mathbf{P}$ and $\mathbf{Q}$, and then fusing the hybrid features of above detected points again to further identify the overlap region between $\mathbf{P}$ and $\mathbf{Q}$ (Sec. \ref{sec:pic});

$\bullet$ SVD for transformation: based on the predicted high-confidence overlap region between the two point clouds, computing final rigid transformation parameters via a differentiable SVD module.

\subsection{Triple features extraction}\label{sec:fe}
We first embed the input point cloud pair and the image into their respective feature spaces, to obtain point-wise or pixel-wise features. Observing that many existing work achieves substantial results on some 3D-related tasks, e.g. normal estimation and depth prediction \cite{qi2018geonet,saha2021learning}, we think certain 3D-aware geometry information also exists in the image space. Inspired by \cite{liu20213d} and \cite{xu2021image2point}, our network generates a simulated 3D feature to provide additional information. Such features are utilized with other features to collaboratively classify whether a point can be projected to the image space and whether a point is within the intersection part of the two input point clouds. 

Specifically, we mimic the 3D feature $f_{3D}^{\mathbf{I}}$, by a small sub-module, which consists of a convolutional layer, a batch normalization layer, and an additional convolutional layer. Based on the corresponding point cloud $\mathbf{P}_{\mathbf{I}}$, the real 3D feature $f_{3D}^{\mathbf{P}_{\mathbf{I}}}$ is obtained via the 3D encoder and an extra batch normalization layer. The two batch normalization layers in 2D encoder and 3D encoder are used to roughly unify the feature distribution of two modalities. A feature loss is formulated to constrain the generation of $f_{3D}^{\mathbf{I}}$. We explain how to compute it in Sec. \ref{sec:loss}. Since $f_{3D}^{\mathbf{P}_{\mathbf{I}}}$ is only involved in calculating the feature loss, $\mathbf{P}_\mathbf{I}$ is required in the training phase, testing unnecessary. This is consistent with our design intuition: only using a single readily-available image to enhance the registration performance. Note that our default implementation uses PointNet++\cite{qi2017pointnet++} for 3D encoder, and PSPNet \cite{zhao2017pyramid} for 2D encoder. These two encoders can be replaced with other state-of-the-art models. 

\subsection{Cross-modality feature fusion}\label{sec:ff}
Since the first stage of the classification is to determine the points that can be projected to the image plane, the information from different modalities should be taken into consideration. To this end, we introduce a cross-modality feature fusion module to effectively mix the above triple deep features. We take the mixture of $\mathbf{P}$ and $\mathbf{I}$ as an example. The input to this module consists of three parts: $f_{3D}^{\mathbf{P}}$,  $f_{3D}^{\mathbf{I}}$, and $f_{2D}^{\mathbf{I}}$. The output is hierarchically computed by: 
\begin{equation}
\begin{aligned}
&
f_{cm1}^{\mathbf{P}} = f_{3D}^{\mathbf{P}} +  \operatorname{MLP}\left(\operatorname{cat}\left[f_{3D}^{\mathbf{P}}, \operatorname{att}(f_{3D}^{\mathbf{P}}, f_{2D}^{\mathbf{I}}, f_{2D}^{\mathbf{I}})\right]\right)\\
&
f_{cm}^{\mathbf{P}} = f_{cm1}^{\mathbf{P}} + \operatorname{MLP}\left(\operatorname{cat}\left[f_{cm1}^{\mathbf{P}}, \operatorname{att}(f_{cm1}^{\mathbf{P}}, f_{3D}^{\mathbf{I}}, f_{3D}^{\mathbf{I}})\right]\right),
\end{aligned}\label{equ:fusion}
\end{equation}
where $\operatorname{MLP}(\cdot)$ denotes a three-layer fully connected network, $\operatorname{cat}(\cdot, \cdot)$ is concatenation, and $\operatorname{att}(\cdot, \cdot, \cdot)$ means the attention model, which weights the image feature using learned weights. We reshape the 2D image feature as $f_{2D}^{\mathbf{I}} \in \mathbb{R}^{(W_1 \cdot H_1)\times C}$ before feeding it to the fusion module. Similarly, we can obtain the fused feature $f_{cm}^{\mathbf{Q}}$. Detailed model structure is illustrated in Figure \ref{fig:cm_module2}.

\subsection{Two-stage classification}\label{sec:pic}
We conducted a statistical analysis on the Bundlefusion dataset \cite{dai2017bundlefusion} over 800 point cloud-image-point cloud triplets, with a 40\% overlap ratio between two point clouds. We observed that about 80\% of points of the overlapping area between two input point clouds are also located in the intermediate image space. Moreover, the image data in Bundlefusion dataset was captured close to the target, with a small resolution of $640\times480$. If we use some other photo-taking devices, e.g. our cellphone, we will capture a larger intermediate image that contains more overlapping points. Hence, instead of directly learning the overlap parts from point clouds, we adopt a two-stage coarse-to-fine classification strategy to detect the overlap region between the point cloud pairs. First, the fused feature $f_{cm}^{\mathbf{P}}$ (or $f_{cm}^{\mathbf{Q}}$) is fed into a classifier, which is composed of a two-layer MLP and a Softmax layer, to compute the probability that each point can be projected into the image, or not. So we can obtain the subset $\mathbf{S}_\mathbf{I}^\mathbf{P}$ (or $\mathbf{S}_\mathbf{I}^\mathbf{Q}$), which are very likely to be projected to the image $\mathbf{I}$ (see from Figure \ref{fig:example} (e) and (g)). Then, the features of the points from $\mathbf{S}_\mathbf{I}^\mathbf{P}$ and $\mathbf{S}_\mathbf{I}^\mathbf{Q}$ are fused again to further segment the potential overlapping regions $\mathbf{S}_\mathbf{Q}^\mathbf{P}$ and $\mathbf{S}_\mathbf{P}^\mathbf{Q}$. The two subsets have similar shapes, but slightly different in point resolution and distribution (see from Figure \ref{fig:example} (f) and (h)). The second feature fusion module is similar to that used in the first classification stage, but we only need to fuse once, namely use the half (left or right) part of Figure \ref{fig:cm_module2}. 

\subsection{SVD for transformation}\label{sec:svd}
In this section, we calculate the transformation parameters $\mathbf{R}$ and $\mathbf{t}$. In the early test, we provided the features of points from the overlapping area to an MLP to regress a 7D vector (a 3D translation and a 4D quaternion) as the rigid transformation. However, we found that the transformation parameters are difficult to regress, even if we design various losses \cite{xiang2017posecnn,xu2021omnet} to supervise them. Inspired by \cite{wang2019deep}, we believe that establishing suitable point-to-point correspondences is more important for solving transformation parameters, while a general approximator, e.g., MLP has poor interpretability and poor stability. Specifically, we establish a soft corresponding relation matrix among the point in $\mathbf{S}_\mathbf{Q}^\mathbf{P}$ and $\mathbf{S}_\mathbf{P}^\mathbf{Q}$, according to their similarities in the embedded feature space. The weighted SVD is then used to solve for the rigid transformation, which has been shown to be differentiable in \cite{wang2019deep}.

\subsection{Joint loss function}\label{sec:loss}
The proposed network is trained end-to-end, using multiple losses w.r.t. the real 3D feature  $f_{3D}^{\mathbf{P}_\mathbf{I}}$ from $\mathbf{P}_\mathbf{I}$, the ground truth classification labels for points in image (denoted as $\hat{l}(\mathbf{S}^\mathbf{P}_\mathbf{I})$ and $\hat{l}(\mathbf{S}^\mathbf{Q}_\mathbf{I})$) and points in another point cloud (denoted as $\hat{l}(\mathbf{S}^\mathbf{P}_\mathbf{Q})$ and $\hat{l}(\mathbf{S}^\mathbf{Q}_\mathbf{P})$), and the ground-truth rigid transformation ($\hat{\mathbf{R}}$ and $\hat{\mathbf{t}}$) as supervisions. 

\paragraph{Feature loss}
To supervise the generation of $f_{3D}^{\mathbf{I}}$ in the 2D network, we use an $L_2$ loss:
\begin{equation}
L_F = ||\mathbb{T}(f_{3D}^{\mathbf{P}_{\mathbf{I}}}) - f_{3D}^{\mathbf{I}}||_2^2,
\end{equation}
where $\mathbb{T}(\cdot)$ means the operator of projecting the real 3D feature into 2D domain. As the dimension of $f_{3D}^{\mathbf{I}}$ is $W_1 \times H_1 \times C$ and  $f_{3D}^{\mathbf{P}_{\mathbf{I}}}$ is $N \times C$, we cannot directly compute their differences. For each point in $\mathbf{P}_{\mathbf{I}}$, we project it to the 2D image space via the camera intrinsic parameters, and treat the nearest pixel as its corresponding pixel.

\paragraph{Classification loss} The goal of two-stage classification is to progressively detect the overlapping region between the input two point clouds $\mathbf{P}$, $\mathbf{Q}$, via an intermediate image $\mathbf{I}$. Therefore, we formulate four constraint terms:
\begin{equation}
\begin{aligned}
L_C = &\operatorname{CE}(\hat{l}(\mathbf{S}^\mathbf{P}_\mathbf{I}), l({\mathbf{S}^\mathbf{P}_\mathbf{I}}))+ \operatorname{CE}(\hat{l}(\mathbf{S}^\mathbf{Q}_\mathbf{I}), l({\mathbf{S}^\mathbf{Q}_\mathbf{I}}))+ \\
&\operatorname{CE}(\hat{l}(\mathbf{S}^\mathbf{P}_\mathbf{Q}), l({\mathbf{S}^\mathbf{P}_\mathbf{Q}}))+\operatorname{CE}(\hat{l}(\mathbf{S}^\mathbf{Q}_\mathbf{P}), l({\mathbf{S}^\mathbf{Q}_\mathbf{P}})),
\end{aligned}
\end{equation}
where $\operatorname{CE}(,\cdot,)$ denotes cross entropy loss and $l(\cdot)$ means the predicted point label.

\paragraph{Transformation Loss} In our early testing, we try to represent both the ground-truth and predicted rotation matrices in the format of quaternion, and calculate their difference, by $L_2$ loss, as follows:
\begin{equation}\label{eq:T1}
L_{T1} = ||\mathbf{q}-\hat{\mathbf{q}}||_2^2+\lambda||\mathbf{t}-\hat{\mathbf{t}}||_{2}^2,
\end{equation}
where $\hat{\mathbf{q}}$ and $\hat{\mathbf{t}}$ are ground-truth quaternion and translation, $\mathbf{q}$ and $\mathbf{t}$ are the predicted results. However, we found the results is less pleasing. For a neural network, it may be hard to control the rotation accuracy only by relying on four parameters. Inspired by \cite{xiang2017posecnn}, we indirectly compute the pose loss to supervise the rotation parameters, as follows:
\begin{equation}
L_{T2} = \sum_{i=1}^{K}||\hat{\mathbf{R}}\mathbf{s}_i - \mathbf{R}\mathbf{s}_i||_2^2 + \lambda||\hat{\mathbf{t}} - \mathbf{t}||_{2}^2, 
\end{equation}
where $\mathbf{s}_i$ is the ground truth overlap point between the two input point clouds, and $K$ is the number of overlapping points. Finally, the total loss is formulated as:
\begin{equation}
L = L_F + \alpha L_C + \beta L_{T2}.
\end{equation}
Note that we compute two pairs of transformation parameters during training, while using the one with less error in testing stage. The entire network can be run in an iterative manner, so the loss is accumulated over iterations. We experimentally find that setting $\alpha$, $\beta$, and $\lambda$ as 1 works well.
\label{sec:method}

\section{Experiments}
\begin{table}[tb]
\scriptsize
\setlength{\tabcolsep}{1mm}
\caption{Quantitative comparison of different methods on three datasets.}
	\begin{tabular}{llcccccc}
		\hline
		& Method   & RMSE($\mathbf{R}$) & MAE($\mathbf{R}$) & RMSE($\mathbf{t}$) & MAE($\mathbf{t}$) & MIE($\mathbf{R}$) & MIE($\mathbf{t}$) \\ \hline
		\multirow{5}{*}{BundleFusion}   & DCP-v2   & 28.61          & 7.56          & 0.76          & 0.53          & 26.99          & 1.53          \\
		& PREDATOR & 14.87          & 4.34          & 0.59          & 0.43          & 14.92          & 1.03          \\
		& OMNet    & 15.09          & 4.89          & 0.65          & 0.57          & 12.74          & 1.17          \\
		& FMR      & 23.62          & 7.91          & 1.01          & 0.70          & 20.96          & 1.98          \\
		& Ours     & \textbf{13.21} & \textbf{3.31} & \textbf{0.51} & \textbf{0.39} & \textbf{10.80} & \textbf{0.94} \\ \hline
		\multirow{5}{*}{KITTI Odemetry} & DCP-v2   & 39.22          & 14.97         & 2.47          & 1.90          & 34.50          & 4.02          \\
		& PREDATOR & \textbf{13.69}          &  \textbf{4.30} & \textbf{1.11}         &   \textbf{0.97}       & \textbf{10.13}         & \textbf{2.06}         \\
		& OMNet    & 15.54          & 5.25          & 1.89          & 1.38          & 13.60          & 3.99          \\
		& FMR      & 21.11          & 8.38          & 2.27          & 1.87          & 19.29          & 4.25          \\
		& Ours     & 16.34 &    6.71     & 1.66  & 1.59  & 10.33 & 3.33          \\ \hline
		\multirow{5}{*}{LiDAR\_Ours}    & DCP-v2   & 17.01          & 6.31          & 0.45          & 0.34          & 16.09          & 1.01          \\
		& PREDATOR & 6.45           & 3.01          &  \textbf{0.17}        & 0.13         & 6.79           & 0.60 \\
		& OMNet    & 6.91           & 2.90          & 0.20          & 0.17          & 6.55           & 0.58          \\
		& FMR      & 12.87          & 4.77          & 0.59          & 0.51          & 10.21          & 1.18          \\
		& Ours   &  \textbf{6.40}  & \textbf{2.98} & 0.19  & \textbf{0.12} & \textbf{5.80}  & \textbf{0.39} \\ \hline
	\end{tabular}
	\label{tab:metrics}
\end{table}

We evaluate our ImLoveNet on three datasets with different kinds of 3D data: 1) BundleFusion dataset \cite{dai2017bundlefusion}, which is an indoor scene RGB-D benchmark; 2) KITTI Odometry dataset \cite{geiger2013vision}, which is an outdoor scene LiDAR benchmark; 3) a small indoor scene dataset acquired by a commercial LiDAR-based scanner, which is built by ourselves. 


\subsection{Dataset and implementation details}\label{sec:dataset}
\paragraph{BundleFusion} The dataset was captured using a depth sensor coupled with an iPad color camera, and consists of 7 large indoor scenes (60m average trajectory length, 5833 average number of frames). Each sequence contains continuous color images and depth maps with a resolution of 640 × 480, as well as the camera's intrinsic and extrinsic parameters. The corresponding point clouds can be reconstructed by the depth maps and the camera intrinsic parameters. The ground-truth rotation and registration can be computed via camera extrinsic parameters. We choose 0-3 sequences for training, 4 for validation, and 5-6 for testing. In training and inference, detailed point cloud-image-point cloud triplets are formed as follows: 1) Randomly select a frame $i$ and the corresponding point cloud is $\mathbf{P}$. 2) The point cloud in the frame $i+100$ is then regarded as $\mathbf{Q}$; We set the frame interval as 100, according to the overlap ratio. The overlap ratio is computed by the Eq. 1 in \cite{huang2021predator} and the mean overlap ratio is around 30\% with the distance threshold equal to 0.05. 3) Choose the image in the frame $i+50$ as the input intermediate image $\mathbf{I}$. Also, the point cloud $\mathbf{P}_\mathbf{I}$ in frame $i+50$ is used in the training phase; 4) The supervision labels are easily computed by the given camera's intrinsic and extrinsic parameters. Finally, we generate 800 pairs for each sequence, with totally 3200 triplets for training, 800 triplets for validation, and 1000 triplets for testing.

\paragraph{KITTI Odometry} Point clouds in this dataset are directly acquired from a 3D Lidar. There are 11 sequences (00-10) with ground-truth trajectories. We observe that: i) most of the points are located on the ground, while sparse above the ground; ii) the intermediate image should be behind of two point cloud frames, for achieving a high overlap between the point cloud and the image; and iii) point cloud-image-point cloud triplets with rich features, such as static cars or buildings, can provide more reliable information. Hence, we do some pre-processing steps (see from 
the supplemental file) and select 1000 triplets for training and 100 for testing.

\paragraph{LiDAR\_Ours} This dataset was constructed by ourselves, which contains 3 indoor scenes. We build it by using a commercial LiDAR-based 3D scanner, i.e., Leica ScanStation P20 with the precision of 3 mm@50 m. This scanner captures the 3D data of a large-scale scene station by station. Point clouds from different stations are then registered together via at least three pairs of static markers. We cropped 40 pairs of point clouds, and took intermediate photos with our cellphone. The overlap ratio is also around 30\%. We use this dataset only for testing.

\paragraph{Implementation details} We run our network 3 iterations during both training and test. The 3D encoder is shared by different input point clouds within each iteration. All sub-modules are shared during iterations. The input point clouds are downsampled and the size is fixed to 6000 ($M=N$). The feature channel $C$ is 256. The leveraged 2D feature map size is $1/8$ of the input image, namely $W_1 = 1/8W$ and $H_1 = 1/8H$. Our network is implemented on Pytorch and is trained for 200 epochs with the Adam optimizer \cite{KingmaB14}. The initial learning rate is $10^{-5}$ and is multiplied by 0.9 every two epochs. The network is trained with BundleFusion and KITTI individually.

\subsection{Competitors}\label{sec:competitors}
Since we focus more on registering low-overlap point clouds, two most advanced and related methods, OMNet \cite{xu2021omnet} and PREDATOR \cite{huang2021predator}, are chosen as competitors. We also compared with the supervised version of FMR \cite{huang2020feature}.  Beside, considering that DCP-v2 \cite{wang2019deep} is better than many ICP-like traditional techniques, we here only compare with DCP-v2, without comparing to those traditional methods. We retrain all compared networks for fair comparison. 

\paragraph{Evaluation metrics} Thanks for the released codes of OMNet \cite{xu2021omnet} and DCP \cite{wang2019deep}, we carefully checked their respective evaluation metrics. Although they use several of the same metrics, they evaluate on different targets. DCP evaluates on the registered point clouds, while OMNet on the Euler angles. To avoid unnecessary misunderstanding, we specify that our metrics are the same as OMNet. These used metrics are root mean squared error ($\operatorname{RMSE}$) and mean absolute error ($\operatorname{MAE}$), and mean isotropic error ($\operatorname{MIE}$) (Please refer to \cite{xu2021omnet} for more details). The smaller these metric values, the better the results.

\subsection{Evaluation on three datasets }\label{sec:bundle}
We quantitatively evaluate the effectiveness of our method on three different datasets, in which the point clouds are generated by different sensors. Table \ref{tab:metrics} is the predicted rotation and translation errors of different methods. As observed, since DCP-v2 and FMR do not pay more attention on the low-overlap region, their results are less pleasing. Our method achieves better results on BundleFusion and LiDAR\_Ours, while performing less better than PREDATIOR and OMNet on KITTI Odometry. The main reason is that PREDATOR does not directly learn the transformation parameters, but employs the RANSAC scheme to compute them based on the learned overlapping probabilities. This scheme is more stable when dealing with sparse and noisy KITTI data.

\begin{table}[tb]
	\scriptsize
	\centering
	\caption{Comparison of registration errors on the BundleFusion dataset of different ablative settings.}
	\begin{tabular}{lcccccc}
		\hline
		Model           & RMSE($\mathbf{R}$) & MAE($\mathbf{R}$) & RMSE($\mathbf{t}$) & MAE($\mathbf{t}$) & MIE($\mathbf{R}$) & MIE($\mathbf{t}$) \\ \hline
		B               & 18.99    & 5.50   & 1.19    & 1.02   & 11.02     & 2.04     \\
		B+I             & 23.22    & 6.13   & 1.34    & 1.21   & 15.78     & 2.37     \\
		B+I+MF          & 20.62    & 4.89   & 1.01    & 0.88   & 12.98     & 1.88     \\
		B+I+MF+CF       & 16.79    & 4.47   & 0.81    & 0.74   & 12.01     & 1.68     \\
		B+I+MF+CF+TC    & 14.08    & 4.10   & 0.77    & 0.62   & 11.26     & 1.04     \\
		B+I+MF+CF+TC+PL & 13.21    & 3.31   & 0.63    & 0.39   & 10.80     & 0.94   \\ \hline 
	\end{tabular}
	\label{tab:ablation}
\end{table}

\begin{figure}[tb]
	\centering
	\includegraphics[width=42mm]{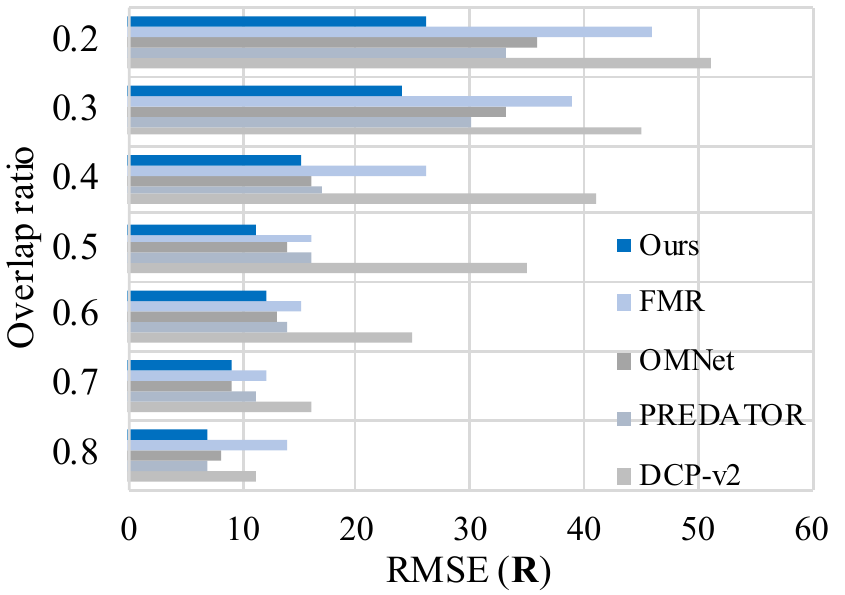} 
	\includegraphics[width=42mm]{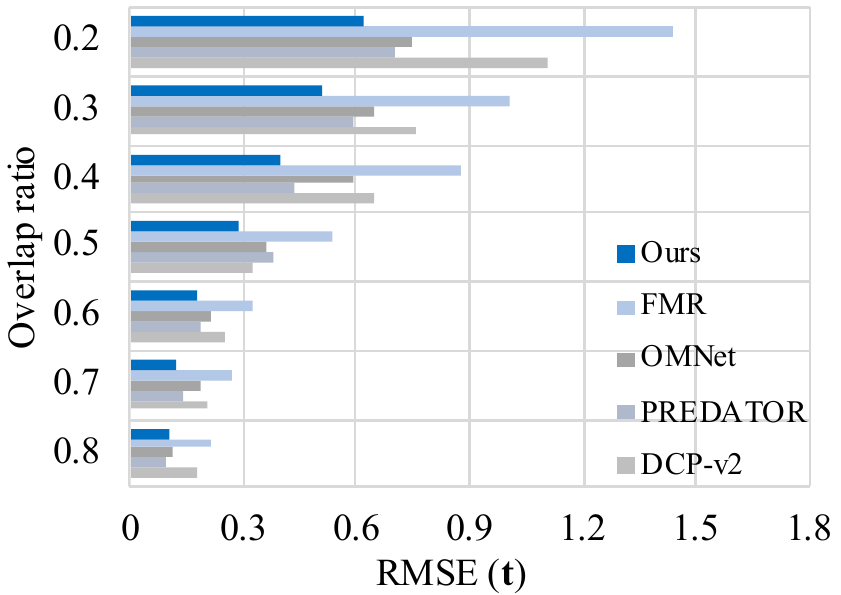}
	\caption{Quantitative comparison of different methods with 7 varying overlap ratios.}
	\label{fig:overlap_ratio}
\end{figure}

\begin{figure}[tb]
	\centering
	\includegraphics[width=0.5\textwidth]{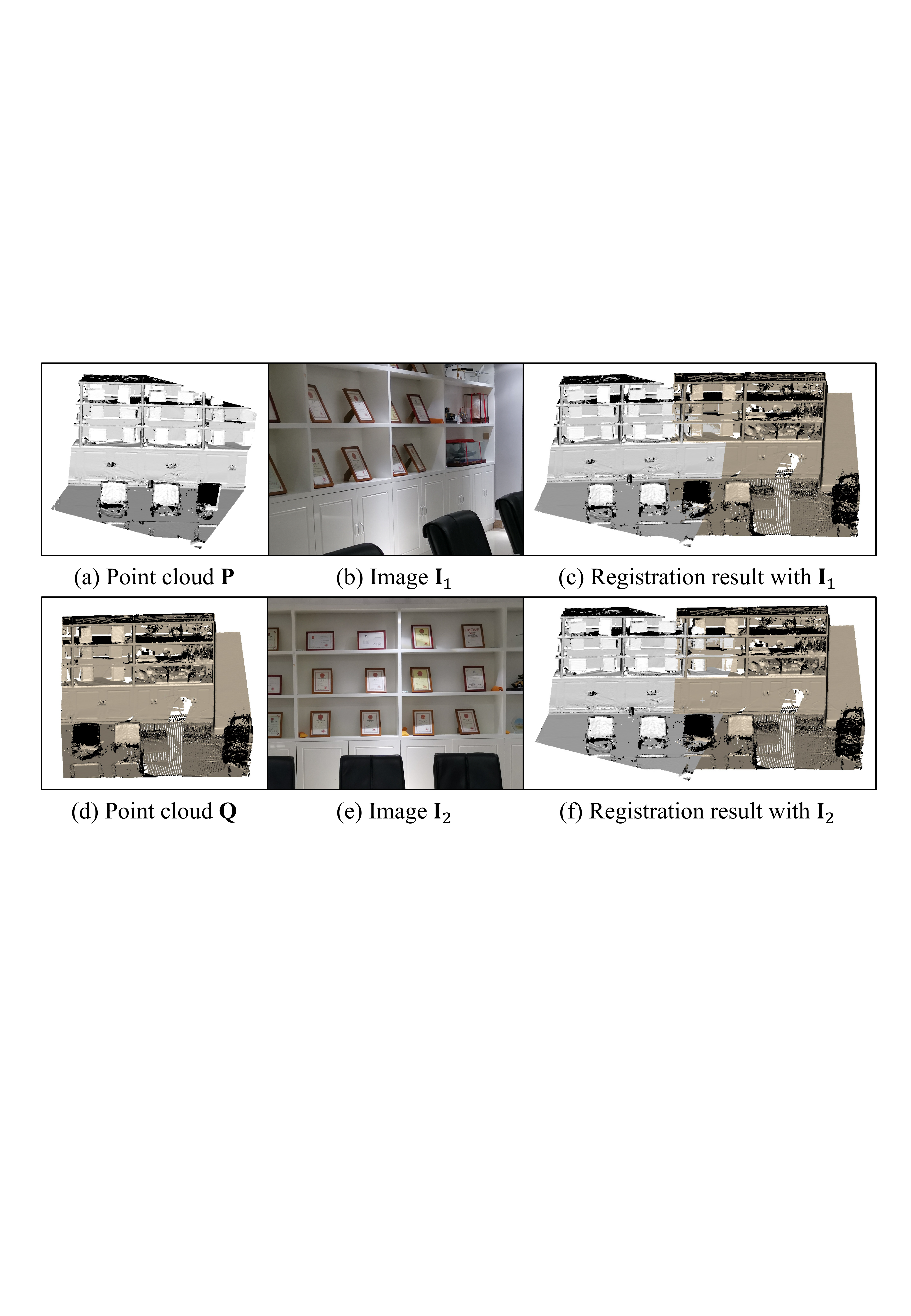}%
	\caption{Registration with two different input images captured from different views and positions. The testing data is from LiDAR\_ours. Although the two images (b) and (e) are different, they both contain the majority part of the overlapping region between the two input point clouds ((a) and (d)).}
	\label{fig:image_view}
\end{figure}

\begin{figure*}[tb]
	\centering
	\includegraphics[width=\textwidth]{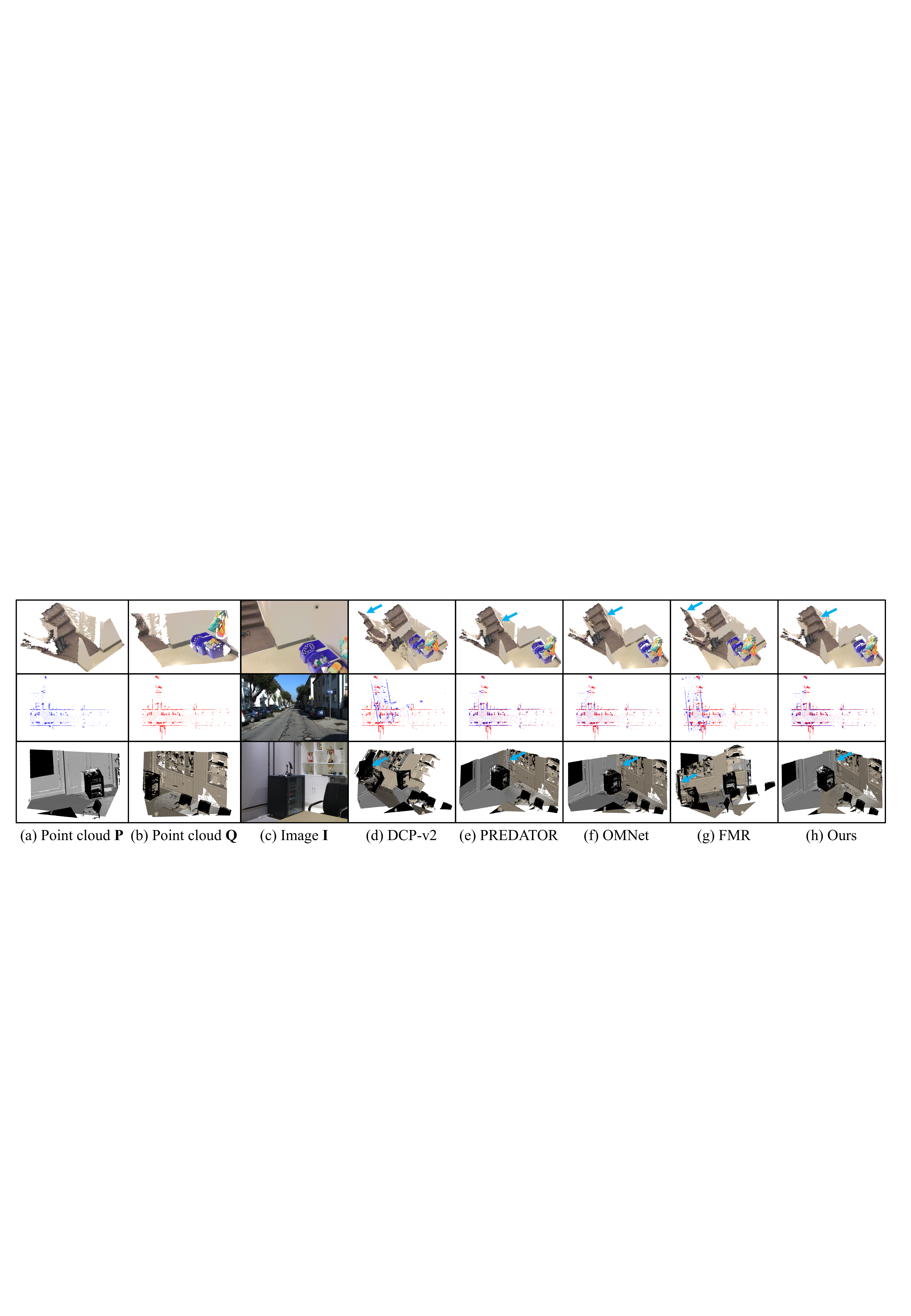}%
	\caption{Visual comparisons of registration results of different methods. The inputs are from BundleFusion, KITTI Odometry, and LiDAR\_Ours. The overlap ratios are set around 30\%. Our method yields significantly better results than DCP-v2 and FMR, and achieves comparable visual results with OMNet and PREDATOR. Blue arrows indicate some registration details for better observation.}
	\label{fig:vis}
\end{figure*}

\subsection{Analysis}\label{sec:analysis}
\paragraph{Ablation study} We ablate the following five main contributions: i) the intermediate image input (\textbf{I}), ii) the mimicked 3D feature (\textbf{MF}), iii) the cross-modality feature fusion (\textbf{CF}); iv) the two-stage classification (\textbf{TC}), and v) the pose loss (\textbf{PL}). Detailedly, if we do not use \textbf{CF}, we replace it with concatenation after necessary projection operation; if we do not use \textbf{TC}, we directly classify overlapping points; if we do not use \textbf{PL}, we use the loss $L_{T1}$ in Eq. \ref{eq:T1}. Table \ref{tab:ablation} clearly reports the contribution of each module on the BundleFusion dataset. Interestingly, we found that directly using image information without any design significantly worsens the final result. This shows the necessity of our subsequent network design.

\paragraph{Classification accuracy} We perform overlapping classification evaluation on the selected BundleFusion dataset and KITTI Odometry dataset. The $precision$ for point-in-image detection is 94\% (BundleFusion) and 90\% (KITTI Odometry), while 81\% and 74\% for the overlap region detection on two point clouds, which means that there are sufficient points to solve the transformation parameters (see visual detection results from the supplemental file).

\paragraph{Different overlap ratios} We evaluate the performance of our model on different overlap ratios. We additional extracted 1,000 test triplets from the BundleFusion dataset. By randomly rejecting the points within the point cloud-point cloud overlap region, we obtain 7 groups of test sets, with varying overlap ratios from 20\% to 80\%. Figure \ref{fig:overlap_ratio} shows that the distributions of $\operatorname{RMSE}(\mathbf{R})$ and $\operatorname{RMSE}(\mathbf{t})$ for different approaches. As observed, when the overlap ratio is lower, our network produces more stable results.

\paragraph{Different image positions} We testify our method with the input image of different image positions and views, as shown in Figure \ref{fig:image_view}. The testing data is from LiDAR\_ours and the used model is trained on the BundleFusion. Although $\mathbf{I}_1$ and $\mathbf{I}_2$ (Figure \ref{fig:image_view} (b) and (e) are captured in different views, they both contain most of the overlapping area between the two input point clouds (Figure \ref{fig:image_view} (a) and (d)). As a result, they both contribute to the registration results. Moreover, we select six kinds of image frame as the intermediate image: one is near the source point cloud ($\textbf{I}_1$), three are very close to the intermediate frame ($\textbf{I}_2$, $\textbf{I}_3$, and $\textbf{I}_4$), one is near the target point cloud ($\textbf{I}_5$), and the last one is a randomly-selected image ($\textbf{I}_6$). We collect twenty groups of data from BundleFusion. Quantitative statistics reported in Table \ref{tab:imagepos} show that if the intermediate image contains a higher overlap rate with both two input point clouds, it will assist to produce a better result.

\paragraph{Effect of image for other registration model} In fact, it is very interesting to see whether the input image contributes to other registration models. This is also meaningful to clarify the contribution of our work. In this section, we try to integrate the image information, as well as the simulated 3D feature into the PREDATOR \cite{huang2021predator}. Specifically, we feed an extra image into PREDATOR, and fuse the extracted image deep features in the encoding step of PREDATOR. Hence, we can obtain two variants: i) only incorporate image information; 2) incorporate both image information and the simulated 3D feature. We compare the performance of the original PREDATOR model and its two variants on the BundleFusion dataset, as shown in Table. \ref{tab:predator}. We observe that directly encoding the image information into the network may not produce better results, while using the simulated 3D feature can help to yield more accurate registration results. Figure \ref{fig:teaser} also shows the visualization results.

\begin{table}
	\footnotesize
	\centering
	\caption{Comparison of registration errors by using different intermediate images.}
	\begin{tabular}{ccccccc}
		\hline
			Method  & $\textbf{I}_1$ & $\textbf{I}_2$ & $\textbf{I}_3$ & $\textbf{I}_4$  & $\textbf{I}_5$ & $\textbf{I}_6$\\ \hline
			RMSE($\mathbf{R}$) & 11.77  & 5.58  &  5.46  &  5.73  & 11.74    & 11.88   \\
		   RMSE($\mathbf{t}$) & 1.03  & 0.92  &  0.94  &  0.92   & 0.97    & 1.01 \\ \hline
	\end{tabular}
    \label{tab:imagepos}
\end{table}

\begin{table}
    \footnotesize
	\setlength{\tabcolsep}{0.6mm}
	\centering
	\caption{Performance comparison of original PREDATOR model and its two variants on the BundleFusion dataset.}
	\begin{tabular}{ccccc}
		\hline
		Model           & RMSE($\mathbf{R}$) & MAE($\mathbf{R}$) & RMSE($\mathbf{t}$) & MAE($\mathbf{t}$) \\ \hline
		 PREDATOR & 14.87          & 4.34          & 0.59          & 0.43                 \\
		PREDATOR w/ image            & 14.29    & 5.26   & 0.63    & 0.57      \\ 
		PREDATOR w/ image  and simulated 3D feature          & 13.42    & 4.03   & 0.45    & 0.39   \\ \hline 
	\end{tabular}
	\label{tab:predator}
\end{table}

\begin{table}
	\footnotesize
	\centering
	 \caption{Quantitative comparison between DeepI2P (twice registration) and our network.}
	\begin{tabular}{lcccccc}
		\hline
			Method  & RMSE($\mathbf{R}$) & MAE($\mathbf{R}$) & RMSE($\mathbf{t}$) & MAE($\mathbf{t}$) & MIE($\mathbf{R}$) & MIE($\mathbf{t}$) \\ \hline
		DeepI2P & 34.78    & 14.19   & 1.23    & 0.85   & 43.32     & 1.23     \\
		Ours    & 24    & 8.70   & 0.51    & 0.29   & 20.78     & 0.37    \\ 	\hline
	\end{tabular}
    \label{tab:deepI2p}
\end{table}

\paragraph{Comparison with DeepI2P \cite{li2021deepi2p}} DeepI2P is designed for registering an image to a point cloud, namely computing the extrinsic parameters of the imaging device with respect to the reference frame of the 3D point cloud. Intuitively, we can register $\mathbf{P}$ and $\mathbf{Q}$ to $\mathbf{I}$, respectively. Then, the relative pose between $\mathbf{P}$ and $\mathbf{Q}$ is achieved. We give an objective evaluation between DeepI2P and our model in Table \ref{tab:deepI2p}. The test data is from the selected KITTI Odometry dataset. Our method produces better results, while DeepI2P suffers from the accumulated errors of twice registration. Note that DeepI2P is retrained on the KITTI dataset.

\paragraph{Generalization ability} Indeed, the generalization ability of a network can alleviate the domain gap problem. Our network is able to generalizes smoothly on the LiDAR\_Ours dataset, which is composed of unseen indoor scenes. To further validate it, we attempt to test on the KITTI dataset with the network trained on BundleFusion. Although the scene objects, like cars, buildings, and trees, are never seen by the model, we observed that the registration result is visually satisfactory (see from the supplemental file).

\paragraph{Timing} When the input point cloud size is 6000 and the image resolution is $640\times480$, the inference time of our model is around 5 seconds with 3 iterations. 

\subsection{Visualizations} \label{sec:vis}
We show visual comparisons of registration results in
Figure \ref{fig:vis}. The inputs are from different datasets: BundleFusion, KITTI Odometry, and LiDAR\_Ours. The overlap ratios are all around 30\%. Our method yields significantly better results than DCP-v2 and FMR, and achieves slight improvement than OMNet and PREDATOR.
\label{sec:expe}

\section{Conclusions}
We have introduced, ImLoveNet, a new deep model designed for pairwise registration of low-overlap point clouds, with the assistance of misaligned intermediate images, which can be easily captured via a readily camera device. ImLoveNet tries to fully utilize and collaborate cross-modality information, to faithfully learn the most reliable overlapping regions for robust registration. Compared with common registration models, one obvious limitation of this method is that it should have additional paired point cloud and image as well as the camera intrinsic parameter for training. In the future, we will try to transfer well-trained 3D or 2D models, to further boost the performance of low-overlap point cloud registration.
\label{sec:conc}

\begin{acks}
This work was supported in part by the National Key Research and Development Program of China (No. 2019YFB1707501), National Natural Science Foundation of China (No. 62172218, No. 62032011), and Natural Science Foundation of Jiangsu Province (No. BK20190016). The corresponding author for this paper is Jun Wang.
\end{acks}

\bibliographystyle{ACM-Reference-Format}
\bibliography{main}


\end{document}